\documentclass{article}

\usepackage{PRIMEarxiv}

\usepackage[utf8]{inputenc} 
\usepackage[T1]{fontenc}    
\usepackage{hyperref}       
\usepackage{url}            
\usepackage{booktabs}       
\usepackage{amsfonts}       
\usepackage{nicefrac}       
\usepackage{microtype}      
\usepackage{lipsum}
\usepackage{fancyhdr}       
\usepackage{graphicx}       
\usepackage{cite}
\graphicspath{{media/}}     

\title{MLP-SRGAN: A Single-Dimension Super Resolution GAN using MLP-Mixer}

\author{
  Samir Mitha, Seungho Choe \\
  Electrical, Computer, and Biomedical Eng.  \\
  Toronto Metropolitan University \\
  Toronto, ON, Canada\\
  \texttt{\{samir.mitha, seungho.choe\}@torontomu.ca} \\
   \And
  Pejman Jahbedar Maralani, Alan R. Moody \\
  Department of Medical Imaging \\
  University of Toronto \\
  Toronto, ON, Canada\\
  \texttt{\{Pejman.Maralani, alan.moody\}@sunnybrook.ca} \\
   \AND
  April Khademi\thanks{Keenan Research Center, St. Michael's Hospital, Toronto, ON, CAN}
					\thanks{Institute of Biomedical Engineering, Science and Technology (iBEST), Toronto, ON, CAN} \\
  Electrical, Computer, and Biomedical Eng.  \\
  Toronto Metropolitan University \\
  Toronto, ON, Canada\\
  \texttt{akhademi@torontomu.ca} \\
}

\begin{document}
\maketitle

\begin{abstract}
We propose a novel architecture called MLP-SRGAN, which is a single-dimension Super Resolution Generative Adversarial Network (SRGAN) that utilizes Multi-Layer Perceptron Mixers (MLP-Mixers) along with convolutional layers to upsample in the slice direction. MLP-SRGAN is trained and validated using high resolution (HR) FLAIR MRI from the MSSEG2 challenge dataset.   The method was applied to three multicentre FLAIR datasets (CAIN, ADNI, CCNA) of images with low spatial resolution in the slice dimension to examine performance on held-out (unseen) clinical data. Upsampled results are compared to several state-of-the-art SR networks. For images with high resolution (HR) ground truths, peak-signal-to-noise-ratio (PSNR) and structural similarity index (SSIM) are used to measure upsampling performance. Several new structural, no-reference image quality metrics were proposed to quantify sharpness (edge strength), noise (entropy), and blurriness (low frequency information) in the absence of ground truths. Results show MLP-SRGAN results in sharper edges, less blurring, preserves more texture and fine-anatomical detail, with fewer parameters, faster training/evaluation time, and smaller model size than existing methods. Code for MLP-SRGAN training and inference, data generators, models and no-reference image quality metrics will be available at \url{https://github.com/IAMLAB-Ryerson/MLP-SRGAN}.
\end{abstract}


\section{Introduction}
Fluid attenuation inversion recovery (FLAIR) MRI is used to diagnose neurological disease including dementia and cerebrovascular disease (CVD). 2D FLAIR acquisitions usually have thick slices, which limits direct comparison with other MRI sequences, or longitudinal scans.  Moreover, inputs to deep-learning and registration tools may require specific spatial resolutions. In the past, bilinear and bicubic interpolation methods were used to change the spatial resolution of images. More recently, super resolution (SR) methods have been proposed for natural images that retain photorealism \cite{Ledig:arXiv:2015}.\\ 
\indent Convolutional neural networks (CNN) have been used to learn the mapping between the low- and high-resolution images for SR applications \cite{Dong:arXiv:2015}. The perceptual output of SRCNN surpasses traditional methods such as bilinear and bicubic interpolation. More recently, generative adversarial networks (GAN) for image super-resolution (SRGAN) have been gaining traction as they maintain photo-realism in natural images for 4$\times$ upscaling \cite{Ledig:arXiv:2015}. SRGAN \cite{Ledig:arXiv:2015} aims to recover finer texture details with a perceptual loss that combines adversarial and content losses. Enhanced SRGAN \cite{Wang:arXiv:2019} introduced the Residual-in-Residual Dense Block without batch normalization and relativistic GANs that lets the discriminator predict relative realness. It produced better visual quality with more realistic and natural textures than SRGAN and is current state-of-the-art. SR has also been utilized in medical imaging with similar networks and loss functions \cite{Wang:arXiv:2019} - \cite{Gu:Journal:2020}, which can restore small structures (i.e. septum pellucidum)  \cite{Rudie:arXiv:2022} and better texture detail \cite{Gu:Journal:2020}. \\
%
%
%
\indent Although these methods show promise, photorealism, resolving fine-details and maintaining texture remain problems when upscaling using SR for medical images.  Existing methods can create anatomical inaccuracies, blurring, smoothing, artefacts, etc. This is especially true for FLAIR MRI with thick slices causing notable interpolation errors and blurring between object boundaries when upsampled.  Existing networks also upscale in 2 dimensions, but for FLAIR MRI, we are concerned with upsampling along the (single) slice dimension. Lastly, methods based strictly on CNNs have short-comings, in that CNNs require a lot of data to train and can fail to encode position and orientation information which may be important for retaining texture or fine-details.\\
\indent To overcome these challenges, we propose a novel architecture called MLP-SRGAN, which is a single-dimension SRGAN that utilizes Multi-Layer Perceptron Mixers (MLP-Mixers) along with convolutional layers to upscale FLAIR MRI volumes.  This is the first time an MLP-Mixer is employed in an SR application. Convolution-free networks are gaining attention in computer vision applications to overcome CNN limitations \cite{Tolstikhin:arXiv:2021} \cite{Cazenavette:arXiv:2022}. The MLP-Mixer \cite{Tolstikhin:arXiv:2021} architecture has been purposed as one such solution, and studies show multi-layered perceptrons are enough for visual learning. MLP-Mixer architectures are based entirely on MLPs repeatedly applied across either spatial locations or feature channels. They accept a sequence of linearly projected patches (tokens) that maintains dimensionality. Channel-mixing MLP communicates between channels, and token-mixing MLPs operate on each token independently. CNNs have been shown to have a dependence on spatial location, which can affect the outcome of vision applications \cite{Kayhan_2020_CVPR}. The channel mixing functionality of the MLP-Mixer helps to reduce this spatial dependency introduced from convolutional layers. Combining MLP-Mixers with convolutional layers for vision tasks have been done before \cite{Cazenavette:arXiv:2022}. The benefit of using MLP-Mixers in conjunction with convolutional layers is a reduction of parameters and faster compute time, while still achieving similar performance to CNN-based solutions.\\
%
\indent For FLAIR MRI we complete upsampling along the slice dimension to interpolate thick slices. We propose a novel block known as the Residual MLP-Mixer in Residual Dense Block, which is used in an SRResNet-like architecture to upscale images by 4$\times$ over a single dimension. We also propose a selective downsampling Block, which uses convolutional layers to select relevant pixels from the fully upscaled images and provide intelligent anti-aliasing for a one-dimensional upscale. The selective downsampling block also allows the output of the network to be scaled to the desired resolution. Results are compared to five popular deep learning-based SR methods and bicubic interpolation. Networks were tested on four multicentre FLAIR MRI datasets and results were compared using peak-signal-to-noise-ratio (PSNR), structural similarity index measure (SSIM), and three novel no-reference image metrics based on sharpness, entropy, and low frequency of the discrete wavelet transform.

\section{Methods} \label{sec:methods}
\subsection{Network Architecture}
\begin{figure}[!htb]
\centering
\includegraphics[scale=0.31]{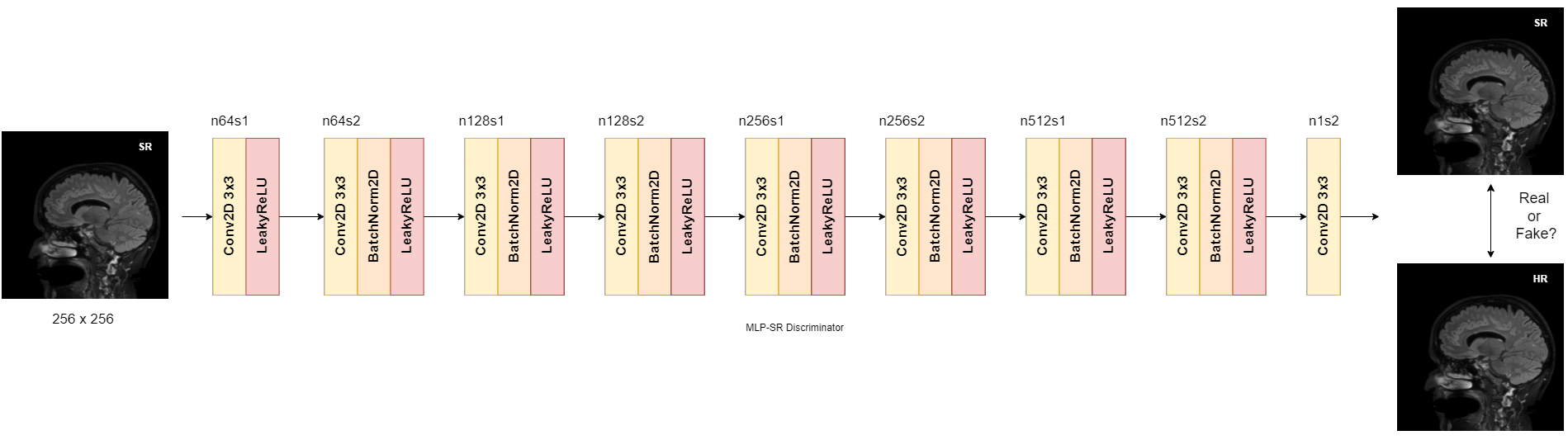}
\vspace{-.2in}\caption{\small MLP-SRGAN discriminator network. $n$ refers to the number of filters in the layer, while $s$ refers to stride size in the layer.}
\label{discriminator_fig}
\end{figure}

%
\indent We propose a new architecture called MLP-SRGAN which contains a combination of MLP-Mixer blocks and convolutions in a generator network (Figure \ref{generator_fig}) and a CNN-based discriminator (Figure \ref{discriminator_fig}). The CNN-based discriminator network allows the generator network to converge faster during training and ensures outputs are photorealistic. The network accepts any input image resolution, and can render any output resolution, which permits us to take advantage of upscaling in a single dimension to address thick slices in FLAIR MRI.
\subsubsection{Generator} 
\begin{figure}[!htb]
\centering
\includegraphics[scale=0.45]{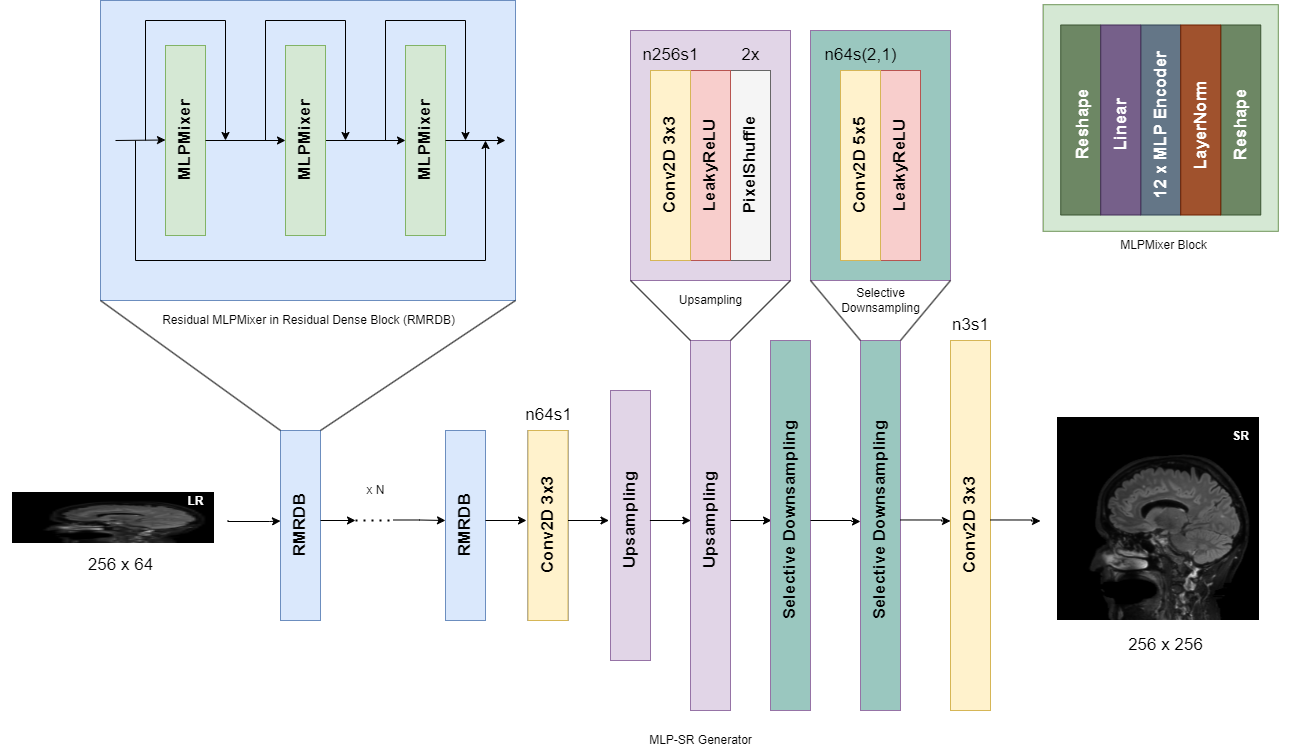}
\vspace{-.2in}\caption{\small MLP-SRGAN generator network. $n$ refers to the number of filters in the layer, while $s$ refers to stride size in the layer.}
\label{generator_fig}
\end{figure}

The generator network consists of MLP-Mixer blocks with a reshaping layer, a linear connecting layer, MLP encoder, followed by layer normalization, and finally another reshaping layer. A residual connection is added from the input of the MLP-Mixer block to the output to ensure information from lower layers is maintained which improves super resolution performance \cite{Wang:arXiv:2019}. Three residual MLP-Mixer blocks are serially connected with a residual connection from the input of the first block to the output of the last block, to create the residual MLP in residual dense block (RMRDB). A varying number of RMRDB are connected together sequentially to form a SRResNet-like architecture, which is connected to convolution blocks and upsampling layers to create a 4$\times$ upscaled image. The 4$\times$ upscaled image is then input into a new block called the selective downsampling block, which provides for a highly flexible mechanism to control the output resolution. \\
\indent The selective downsampling block consists of a convolutional layer followed by leaky ReLU activation as shown in Figure \ref{generator_fig}. By strategically selecting kernel size and stride of the convolutional portion of the selective downsampling block, the output resolution of the network can be intelligently downsampled by an integer value. By changing the kernel size and stride of the downsampling block, and changing the number of upsampling layers, the output resolution of the volume can be scaled to any value. The flexibility of the network allows for upscaling to higher resolutions by increasing the number of RMRDBs, upsampling layers, and selective downsampling layers respectively.\\
\indent For this experiment, kernel size of 5x5 is chosen to ensure information from the upscaled low resolution input is contained within each convolution operation. A stride of 2x1 ensures the convolution output size is scaled down by a factor of 2 in a single dimension for each selective downsampling layer. Using this kernel and stride size allows the network to achieve a 4$\times$ upscale over the slice direction, thereby focusing the model on the thick slices, while still maintaining the original image quality in the other dimensions.
\subsubsection{Discriminator} The discriminator consists of convolutions, batch normalization, and leaky ReLU activations similar to  \cite{Ledig:arXiv:2015}, with the final dense layers and sigmoid activation replaced with a convolution layer to save memory.
\subsection{Loss Functions}
\indent The loss functions and loss function parameters used are similar to those in \cite{Wang:arXiv:2019}.  The generator uses perceptual loss, content loss, and adversarial loss
\begin{equation}
\small L_{gen}=\lambda_1 L_{percep} + \lambda_2 L_{content} + \lambda_3 L_{adv}
\end{equation}
where $\lambda_1=1$, $\lambda_2=0.01$, and $\lambda_3 = 0.005$.  The perceptual loss uses features from the last convolution layer of VGG19. Mean absolute error is computed from features of the high resolution image $I_{HR}$ and generated super resolution image $I_{SR}$ to create a perceptual loss
\begin{equation}
\small  L_{percep}=\sum_{i=1}^{N}\frac{|vgg19(I_{HR})-vgg19(I_{SR})|}{N}
\end{equation}
where $vgg19$ is the VGG19 feature extraction network. The content loss is the mean absolute error between the pixel values of the original high resolution image $I_{HR}$ and the generated super resolution image $I_{SR}$. The adversarial loss for the generator is the binary cross entropy of 1 minus the relativistic average discriminator output, while the discriminator loss is the binary cross entropy of the discriminator output:
\begin{equation}
\small L_{adv}=-\log(1-D(I_{HR},I_{SR}))-\log(D(I_{SR},I_{JR}))
\end{equation}
\begin{equation}
\small L_{discr}=-\log(D(I_{HR},I_{SR}))-\log(1-D(I_{SR},I_{HR}))
\end{equation}
where $D$ is the discriminator network. The generator and discriminator are trained alternately during each iteration of training.
\subsection{Image Quality Metrics}
This section presents the image quality metrics used in this work.  Let $I(x,y)$ represent an image where $(x,y)\in Z^2$ are the spatial coordinates and $L$ is number of graylevels with $I\in[0, L-1]$. The human perceptual metreics are the PSNR, SSIM and the structural image quality metrics are Shannon's Entropy, Sharpness and Wavelet Low (Blurriness).  The equations for each metric are summarized below.
\begin{center}
\begin{tabular}{ |c|c| } 
 \hline
 Name & Metric  \\ \hline
 Peak Signal-to-Noise-Ratio (PSNR) & $ PSNR=10 \cdot \log _{10}\left(\frac{M A X_I^2}{M S E}\right)$\\\hline
 Structural Similarity Index Measure (SSIM) & $SSIM=\frac{\left(2 \mu_{I_1} \mu_{I_2}+c_1\right)\left(2 \sigma_{{I_1}{I_2}}+c_2\right)}{\left(\mu_{I_1}^2+\mu_{I_2}^2+c_1\right)\left(\sigma_{I_1}^2+\sigma_{I_2}^2+c_2\right)}$  \\ 
 \hline
  Shannon's Entropy & $
H(I)=-\sum_{i=0}^{L-1} p\left(I\right) \log p\left(I\right) 
$  \\ \hline
Sharpness & $
IS(I)=\frac{1}{N} \sum_{x,y}H_{sobel}(x,y) * I(x,y) 
$ \\ \hline
Blurriness (Wavelet Low) & $
Low(I)=\sum_{s=1}^{5}\frac{A^{2}_s(x,y)}{N} 
$ \\ \hline
\end{tabular}
\label{t:metrics}
\end{center}
\begin{itemize}
    \item \textbf{PSNR}: $M A X_I$ represents the maximum intensity given the image's bit depth, while $M S E$ is the mean squared error between the generated and ground truth images. 
    \item \textbf{SSIM}: ($\mu_{I_1}$,$\mu_{I_2}$) and ($\sigma_{I_1}^2$, $\sigma_{I_2}^2$) are the mean and variance of the generated and ground truth images. $\sigma_{{I_1}{I_2}}$ is the covariance between generated and ground truth images. $c_1$ and $c_2$ are constants that ensure the metric does not exceed 1.
    \item \textbf{Shannon's Entropy}: $p(I)$ is the probability distribution (normalized number of occurences) of intensity $I$. Entropy measures the randomness or noise levels in an image and a high value indicates more rapid intensity variations.
    \item \textbf{Sharpness} $H_{sobel}(x,y)$ is the 2D sobel filter used for edge detection, $*$ represents the convolution operation, and $N$ is the total number of pixels in the image.  High quality images have high contrast and sharp edges which in turn, have large edge magnitudes and an overall high Sharpness metric. 
    \item \textbf{Blurriness}: the low frequency band (approximation coefficients) of the DWT are specified by $A_s(x,y)$, where $s$ is the scale (decomposition level). Five levels ($s=5$) were used with a Daubechies wavelet.  This metric considers the energy in the low frequency bands. In images with more blurring, there is higher energy in the low frequency approximations, and the Wavelet metric would be high. 
\end{itemize}

\section{Experiments}
\subsection{Data} Training data is from the MSSEG2 challenge \cite{commowick_objective_2018} and consists of 80 high resolution FLAIR volumes of 256$\times$256$\times$256, with 0.9766mm$\times$0.9766mm$\times$0.5-1.0mm resolution from GE, Philips and Siemens scanners. Sixty-four (64) volumes are used for training/validation, and the remainder 16 are used for testing. Volumes are split into individual slices in the sagittal plane with blank slices removed for training, resulting in a total of 10,940 sagittal slices used to train each network. The original high-resolution (HR) slices are used as ground truth, and low-resolution (LR) images are created by downsampling the original images to 256$\times$64 with bicubic interpolation. All models under go a five fold cross validation so all 80 volumes are used for testing purposes without data leakage.  One hundred (100) volumes randomly sampled from three additional multicentre, clinical data sets are used as hold out, unseen datasets. The datasets are from the Canadian Atherosclerosis Imaging Network (CAIN) dataset \cite{Tardif:Journal:2013}, the Canadian Consortium of Neurodegeneration and Aging (CCNA) dataset \cite{mohaddes_national_2018}, and the Alzheimer's Disease Neuroimaging Initiative (ADNI) \cite{jack2008alzheimer}. Acquisition parameters are shown in Table \ref{tab:0}.  All volumes are intensity normalized \cite{Brittany:Journal:2019}.
\begin{table}[!h]
    \small
    \caption{Dataset acquisition parameters. All data is 3T.}
    \label{tab:wml_paper_tab1}
    \centering
    \begin{tabular}{||c|c|c|c|c|c|c||}\hline
    & \multicolumn{6}{|c|}{\textbf{Acquisition Parameters}}\\
     \hline
    \textbf{Database} & \textbf{GE/Phil/Siem} &  \textbf{TR (ms)}  & \textbf{TE (ms)}  & \textbf{TI (ms)}
     & \textbf{XY (mm)} & \textbf{Slice (mm)} \\
    \hline
        MSSEG2  & N/A & N/A & N/A & N/A & 0.9766 & 0.5-1 \\
    \hline
        CAIN  & 27/24/49 & 9000-11000 & 117-150 & 2200-2800 & 0.4285-1 & 3-5 \\
    \hline
        ADNI  & 23/15/62 & 9000-11000 & 90-150 & 2250-2500 & 0.8594 & 2-5 \\
    \hline
        CCNA  & 10/16/74 & 9000-10000 & 115-145 & 2250-2500 & 0.9375 & 3-6 \\
    \hline
    \end{tabular}
    \label{tab:0}
\end{table}
\subsection{Training Details} To fairly compare with other super resolution experiments, the proposed network performs a 4$\times$ upscale. A batch size of 8 is used, with  low resolution inputs sized to 256$\times$64 and output images sized to 256$\times$256. Adam optimizer with default PyTorch parameters, a learning rate of 2e-4 and a decay beginning in the 100th epoch are used. The network uses a pre-trained VGG19 network for the perceptual loss. Since this network requires a 3-channel input image, the input images are duplicated into 3 channels. The VGG19 pre-trained model requires images to be normalized to a specific intensity range and default VGG19 parameters are used $\mu_R = 0.485$, $\mu_G = 0.456$, $\mu_B = 0.406$, and $\sigma_R = 0.229$, $\sigma_G = 0.224$, $\sigma_B = 0.225$. The network interpolation strategy proposed by \cite{Wang:arXiv:2019} is used to train the generator, where the generator network is trained alone for 500 iterations using only the content loss. This strategy of warm up iterations helps to improve stability once the perceptual and adversarial loss when the discriminator is added as the generator is initialized with weights that help to maintain intensity.
\subsection{Image Quality Metrics}  Several image quality metrics are used to evaluate upsampling performance. Peak Signal-to-Noise-Ratio (PSNR) and Structural Similarity Index Measure (SSIM) were used as they traditional metrics used in the literature. We also define three new no-reference based metrics to quantify noise/randomness, sharpness, and blurriness in Appendix A: Table \ref{t:metrics}. The first metric uses Shannon's entropy to measure noise/randomness in the upscaled images - a random image would have higher entropy. The second metric is Sharpness which measures the average magnitude of the image's edge content from the Sobel gradient - an image with high contrast and sharp edges would have a large Sharpness metric. The third metric measures the energy of the low frequency bands from the wavelet decomposition to quantify image blurriness - blurry images have high wavelet energy.
%
%
%
\section{Results}
\begin{figure}[!bht]
\centering
\includegraphics[width=\textwidth]{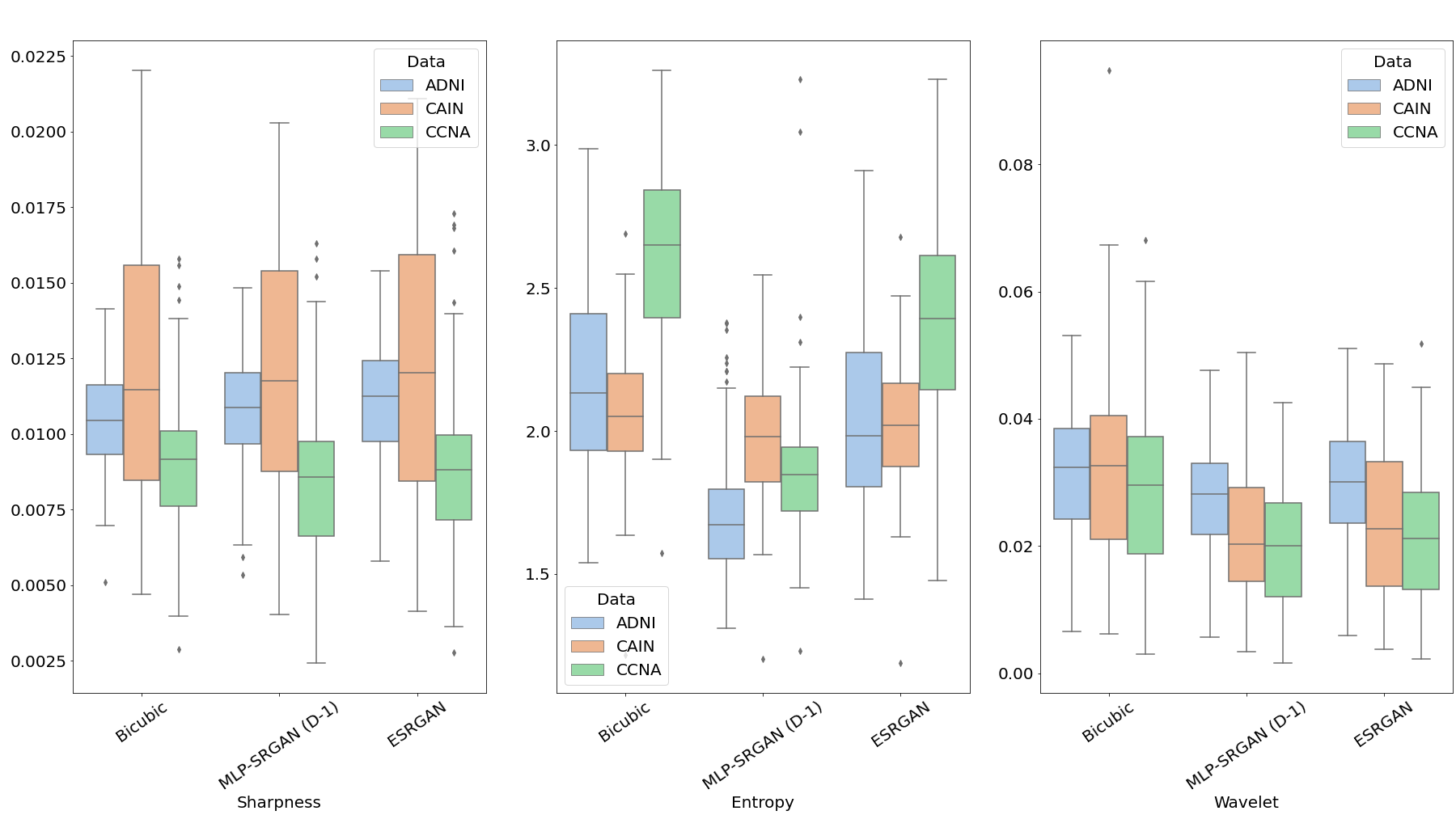}
\caption{Image quality metrics between MLP-SRGAN (D-1) and ESRGAN.}
\label{boxplot_fig}
\end{figure}

\begin{figure}[!ht]
\centering
\includegraphics[width=.73\textwidth]{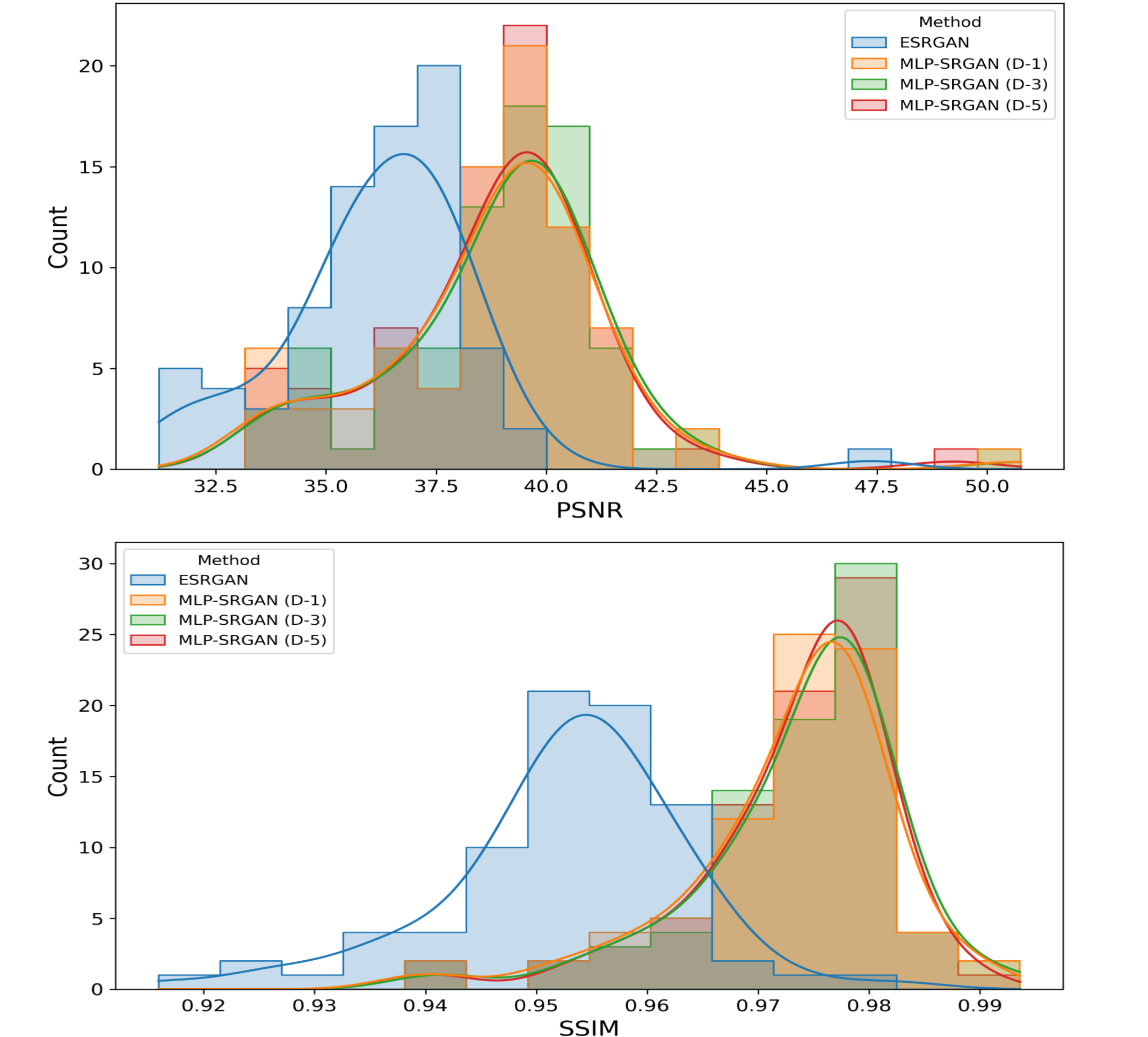}
\caption{Perceptual image quality metrics PSNR and SSIM on MSSEG data.}
\label{ref_line_fig}
\end{figure}

\begin{figure}[!ht]
\centering
\includegraphics[width=.73\textwidth]{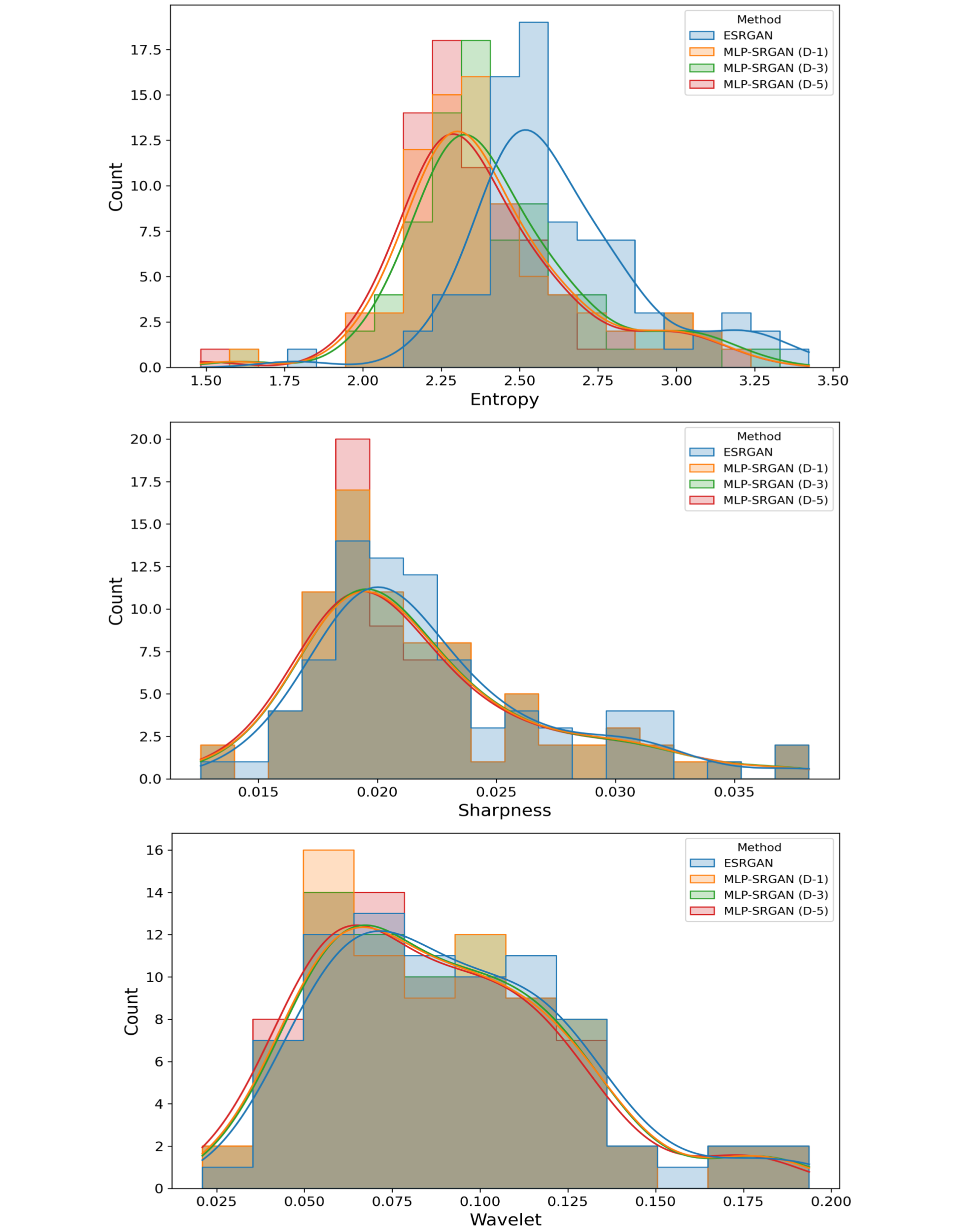}
\caption{Structural image quality metrics Sharpness, Entropy and Wavelet on MSSEG.}
\label{no_ref_line_fig}
\end{figure}

%
Performance of MLP-SRGAN was compared to bicubic interpolation, EDSR \cite{Lim:arXiv:2017}, WDSR \cite{Yu:arXiv:2018}, SRGAN \cite{Ledig:arXiv:2015}, ESRGAN \cite{Wang:arXiv:2019}, and SRCNN \cite{Dong:arXiv:2015}.  MLP-SRGAN was tested with varying number of RMRDBs, represented with (D-N), where N is the number of RMRDBs in the network. MLP-SRGAN (D-1) was also tested without a discriminator, using only the generator and excluding adversarial loss. Training time, evaluation time, model size, and trainable parameters is shown in Table \ref{tab:2}.  MLP-SRGAN has lower training time and approximately same inference time compared to ESRGAN.  Increasing the number of RMRDBs increases the number of parameters, training/inference time. To determine whether performance is different between two models, paired-tests are performed on log-transformed metrics. \\
\begin{table}[ht]
\small
\caption{\small Time and space complexity of networks tested in the five fold cross validation. The MLP-SRGAN has significantly lower training time compared to ESRGAN.}
\centering
  \begin{tabular}{| c | c | c | c | c | c |}
  \hline\hline
    \textbf{Method} & \shortstack{\textbf{Train} \\ \textbf{Time} \\ \textbf{(hh:mm:ss)}} & \shortstack{\textbf{Eval.} \\ \textbf{Time (s)}} & \textbf{Size (MB)} & \shortstack{\textbf{Trainable} \\ \textbf{Params}} & \textbf{GAN}\\
    \hline
    \textbf{Bicubic} & 0 & 0.002 & 0 & 0 & N \\
    \hline
    \textbf{EDSR} & 20:50:22 & 0.006 & 159.9 & 41,900,545 & N \\
    \hline
    \textbf{WDSR} & 3:25:30 & 0.009 & 2.8 & 700,350 & N \\
    \hline
    \textbf{SRCNN} & 1:12:39 & 0.0028 & 0.226 & 57,281 & N \\
    \hline
    \textbf{SRGAN} & 11:11:25 & 0.0048 & 6.0 & 6,244,183 & Y \\
    \hline
    \textbf{ESRGAN} & 52:18:22 & 0.0317 & 147.3 & 43,242,820 & Y \\
    \hline
    \textbf{MLP-SRGAN (D-1)} & 23:17:52 & 0.0144 & 79.9 & 25,595,460 & Y \\
    \hline
    \textbf{MLP-SRGAN (D-3)} & 32:34:22 & 0.0376 & 253.9 & 66,391,236 & Y \\
    \hline
    \textbf{MLP-SRGAN (D-5)} & 42:15:06 & 0.0605 & 392.0 & 107,187,012 & Y \\
    \hline
    \textbf{MLP-SRGAN (D-1) No D} & 18:12:26 & 0.0151 & 79.9 & 20,901,763 & N \\
    \hline\hline
  \end{tabular}
  \label{tab:2}
\end{table}
\begin{figure}[h]
\centering
\includegraphics[scale=0.23]{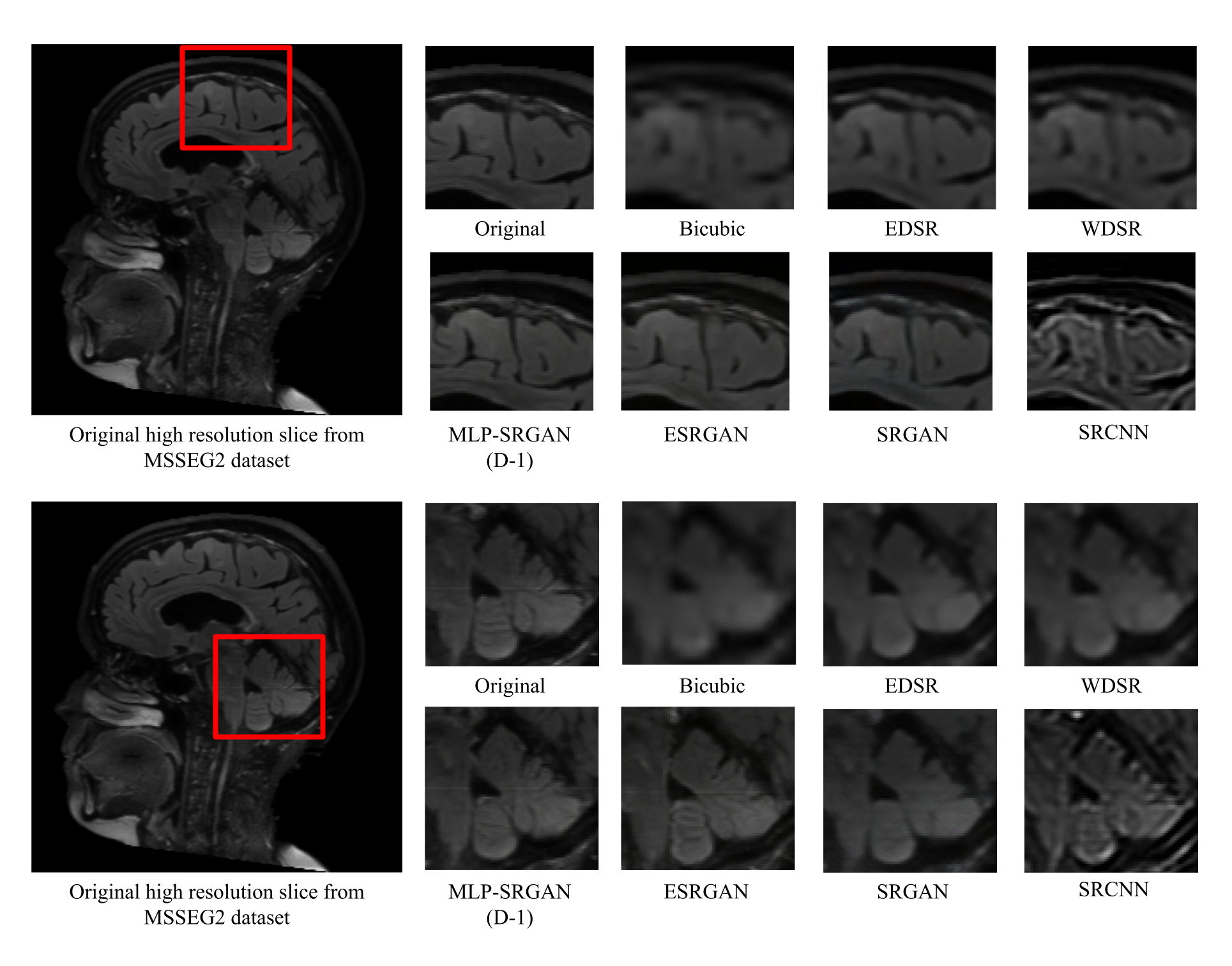}
\vspace{-.23in}\caption{\small Visual comparison of clinically relevant regions from MSSEG2 FLAIR (holdout). }
\label{close_up_fig}
\end{figure}
\indent See Figure \ref{close_up_fig} for generated images from select models in MSSEG test data. Generated images for all methods and datasets are shown in Figure \ref{results2_fig}. Images generated by MLP-SRGAN produce noticeable sharpness in edge content (less blur), with more texture compared to other methods, including ESRGAN. There are differences in small anatomical structures between ESRGAN and MLP-SRGAN. As shown in sulci/gyri (top in Figure \ref{close_up_fig}), the MLP-SRGAN  more accurately represents anatomical details of the original image and has high quality upsampling for fine details. Using MLP-SRGAN, fine details of the cerebellum (bottom) closely follow the original ground truth image, including shape/edges/texture. ESRGAN output shows fine-details are lost, increased blur and there is lower shape and texture correspondence with the original. The high quality generation of fine details, anatomical accuracy, and improved texture of the MLP-Model is retaining small details.\\
\indent Performance metrics for the in-distribution MSSEG2 dataset are shown in Table \ref{tab:1} (average over five folds). The distribution of all the metrics is shown in Figure \ref{ref_box_fig} and \ref{no_ref_box_fig}. The ground truth metrics are computed from the high resolution ground truth images and serve as gold standards for PSNR, SSIM, edge quality (Sharpness), noisiness (Entropy) and blurriness (Wavelet). For perfect upsampling, the metrics from the generated images would be identical to those from the ground truths.  MLP-SRGAN without the discriminator performed the worst (most blur (Wavelet) similar to bicubic, highest randomness (Entropy) and Sharpness most dissimilar). As a result, MLP-SRGAN (No Discr) is not considered further. Images generated by MLP-SRGAN have image quality metrics closer to the gold standard metrics, shown as bold in the table (and red line in the box plots). This shows a discriminator helps to resolve fine-details and improve realness of the images.\\ 
\indent Top models are MLP-SRGAN (D-1) and MLP-SRGAN (D-3). Since MLP-SRGAN (D-1) is more computationally efficient, we choose to further analyze this model and benchmark it against ESRGAN (current state-of-the-art SR method). Metrics from Table \ref{tab:1} show images generated from MLP-SRGAN (D-1) are more similar to ground truth (smallest metric difference) compared to ESRGAN. T-tests were used to compare metrics between MLP-SRGAN and ESRGAN and p-values are shown in Table \ref{tab:MSSEG-p-values}. Over all metrics, there were statistical differences between MLP-SRGAN and ESRGAN ($p<0.05$) for SSIM, PSNR, and Entropy, indicating a significant improvement in image quality as explained by these metrics. These trends are supported by histograms of the metrics in Figure \ref{ref_line_fig} and Figure \ref{no_ref_line_fig}. We hypothesize the fine anatomical details and texture preservation (which we observed visually), combined with sharper edges, less blurriness and more smoothness, contributes to these differences. There were no differences between MLP-SRGAN and ESRGAN with respect to the Sharpness or Wavelet measures (they have overlapping histograms). MLP-SRGAN (D-1) achieves these results with fewer parameters, 2.26$\times$ faster training time, 2.20$\times$ evaluation time, and 0.54$\times$ smaller model size compared to ESRGAN.\\
\begin{table}[!h]
    \small
    \caption{P-values for t-tests comparing MLP-SRGAN and ESRGAN for MSSEG2}
    \label{tab:MSSEG-p-values}
\begin{center}
\begin{tabular}{ |c|c|c|c|c|c| } 
 \hline
                           & PSNR    & SSIM   & Sharpness & Entropy & Wavelet  \\ \hline
MLP-SRGAN (D-1) (No Discr) & $<$0.01       & $<$0.01      & $<$0.01         & $<$0.01       & $<$0.01 \\\hline
MLP-SRGAN (D-1)            & $<$0.01 & $<$0.01      & 0.62   & $<$0.01 & 0.59 \\\hline
MLP-SRGAN (D-3)            & $<$0.01 & $<$0.01      & 0.66      & $<$0.01 & 0.65 \\\hline
MLP-SRGAN (D-5)            & $<$0.01 & $<$0.01      & 0.55   & $<$0.01 & 0.37 \\\hline
\end{tabular}
\end{center}
\end{table}
%
%
%
%
\indent The MLP-SRGAN (D-1), ESRGAN and bicubic interpolation methods are evaluated further on held-out (unseen) clinical datasets (CAIN, ADNI, CCNA). The clinical datasets do not have HR ground truths  and therefore, we cannot compute PSNR or SSIM. Instead, we use the no-reference metrics Entropy, Sharpness and Wavelet and the distributions are shown in Figure \ref{boxplot_fig}. T-tests comparing MLP-SRGAN to ESRGAN are shown in Table \ref{tab:ADNI-p-values} for ADNI, Table \ref{tab:CAIN-p-values} for CAIN and Table \ref{tab:CCNA-p-values} for CCNA.  MLP-SRGAN (D-1) has similar Sharpness to ESRGAN, with lower Entropy and Wavelet metrics. Lower Entropy indicates smoother textures (less noise/randomness) in upscaled images for MLP-SRGAN (D-1).There are significant differences in Sharpness, Entropy and Wavelet between MLP-SRGAN and ESRGAN for the CAIN and CCNA datasets. In ADNI, there are some metrics that were similar, including Wavelet and Sharpness, but differences for Entropy.  Entropy was significantly lower and different in MSSEG as well as  closer to the ground truth on MSSEG, indicating this may be a good metric for future studies on texture differences between images. A lower Wavelet feature for MLP-SRGAN (D-1) indicates lower energy in the low frequency bands (less blurriness) which can be attributed to the intelligent upsampling along the (thick) slice dimension. This was statistically the same as ESRGAN, so blur may be similar. Since these are global metrics it may be hard to quantify local differences. In the future, we will examine metrics that analyze more local features, fine anatomical structures and texture.
\begin{table}[ht]
\small
\caption{Image quality metrics for the MSSEG2 dataset five fold cross validation. Underlined values indicate metrics closest to HR ground truth.}
\centering
  \begin{tabular}{| c | c | c | c | c | c |}
  \hline\hline
    \textbf{Method} & \textbf{PSNR} & \textbf{SSIM} & \textbf{Sharpness} & \textbf{Entropy} & \textbf{Wavelet}\\
    \hline
    \textbf{Ground Truth} & inf & 1.0 & 0.0220 & 2.384 & 0.0874 \\
    \hline
    \textbf{Bicubic} & 36.077 & 0.951 & 0.0168 & 2.462 & 0.1131 \\
    \hline
    \textbf{EDSR} & \textbf{{39.213}} & 0.973 & 0.0194 & 2.404 & 0.1011 \\
    \hline
    \textbf{WDSR} & 38.213 & 0.969 & 0.0193 & 2.441 & 0.1020 \\
    \hline
    \textbf{SRCNN} & 31.304 & 0.915 & 0.0279 & 2.322 & 0.0559 \\
    \hline
    \textbf{SRGAN} & 39.126 & 0.973 & 0.0205 & 2.433 & 0.1010 \\
    \hline
    \textbf{ESRGAN} & 36.241 & 0.953 & 0.0223 & 2.623 & 0.0914 \\
    \hline
    \textbf{MLP-SRGAN (D-1)} & 38.870 & 0.973 & \textbf{{0.0219}} & 2.404 & \textbf{{0.0884}} \\
    \hline
    \textbf{MLP-SRGAN (D-3)} & 38.970 & \textbf{{0.974}} & \textbf{{0.0219}} & 2.429 & 0.0889 \\
    \hline
    \textbf{MLP-SRGAN (D-5)} & 38.868 & 0.973 & 0.0218 & \textbf{{2.390}} & 0.0863 \\
    \hline
    \textbf{MLP-SRGAN (D-1) (No Discr)} & 31.543 & 0.808 & 0.0303 & 3.530 & 0.1108 \\
    \hline\hline
  \end{tabular}
  \label{tab:1}
\end{table}

\section{Conclusion}
We proposed a novel architecture called MLP-SRGAN for upscaling FLAIR MRI images in a single dimension. The proposed method consists of a combination of MLP-Mixers and convolutions in the generator network and convolutions in the discriminator network. The reconfigurable RMRDB blocks, upsampling layers, and selective downsampling layers allow the network to be scaled to higher output resolutions. MLP-SRGAN (D-1) has significantly better performance ($p<0.05$) on  testing sets, faster training and evaluation times compared to state-of-the-art methods such as ESRGAN. Visual analysis shows better retaining of texture, small anatomical details, with less blurring and noise, and higher quality edges in images generated from MLP-SRGAN. We hypothesize the MLP-Mixer blocks are able to retain fine-features by learning mappings from raw image pixels, which have a spatial dependency (compared to CNNs that fail to encode position and orientation information).

\section*{Acknowledgments}
We acknowledge the Natural Sciences and Engineering Research Council (NSERC) of Canada (Discovery Grant), Alzheimer's Society Research Program (ASRP) (New Investigator Grant) and the Ontario Government (Early Researcher Award) for funding this research.

\bibliographystyle{unsrt}  
\bibliography{references}  

\appendix
\newpage
\section{Supplemental Data}
\begin{table}[!h]
    \small
    \caption{P-values for t-tests comparing MLP-SRGAN and ESRGAN for CAIN}
    \label{tab:CAIN-p-values}
\begin{center}
\begin{tabular}{ |c|c|c|c| } 
 \hline
                           & Sharpness & Entropy & Wavelet  \\ \hline
MLP-SRGAN (D-1) (No Discr) & $<$0.01      & $<$0.01    & 0.9 \\\hline
MLP-SRGAN (D-1)            & 0.01      & $<$0.01    & $<$0.01 \\\hline
MLP-SRGAN (D-3)            & 0.05      & $<$0.01    & $<$0.01 \\\hline
MLP-SRGAN (D-5)            & $<$0.01      & $<$0.01    & $<$0.01 \\\hline
\end{tabular}\label{t:metrics}
\end{center}
\end{table}
\begin{table}[!h]
    \small
    \caption{P-values for t-tests comparing MLP-SRGAN and ESRGAN for ADNI}
    \label{tab:ADNI-p-values}
\begin{center}
\begin{tabular}{ |c|c|c|c|c|c| } 
 \hline
                           & Sharpness & Entropy & Wavelet  \\ \hline
MLP-SRGAN (D-1) (No Discr) & 0.27      & $<$0.01    & 0.05 \\\hline
MLP-SRGAN (D-1)            & $<$0.01   & $<$0.01    & 0.41 \\\hline
MLP-SRGAN (D-3)            & 0.36      & $<$0.01    & 0.1  \\\hline
MLP-SRGAN (D-5)            & 0.01      & $<$0.01    & $<$0.01 \\\hline
\end{tabular}\label{t:metrics}
\end{center}
\end{table}
\begin{table}[!h]
    \small
    \caption{P-values for t-tests comparing MLP-SRGAN and ESRGAN for CCNA}
    \label{tab:CCNA-p-values}
\begin{center}
\begin{tabular}{ |c|c|c|c|c|c| } 
 \hline
                           & Sharpness & Entropy & Wavelet  \\ \hline
MLP-SRGAN (D-1) (No Discr) & $<$0.01   & $<$0.01    & 0.04 \\\hline
MLP-SRGAN (D-1)            & $<$0.01      & $<$0.01    & $<$0.01 \\\hline
MLP-SRGAN (D-3)            & $<$0.01      & $<$0.01    & $<$0.01 \\\hline
MLP-SRGAN (D-5)            & $<$0.01      & $<$0.01    & $<$0.01 \\\hline
\end{tabular}\label{t:metrics}
\end{center}
\end{table}

\begin{figure}[!ht]
\centering
\includegraphics[scale=0.15]{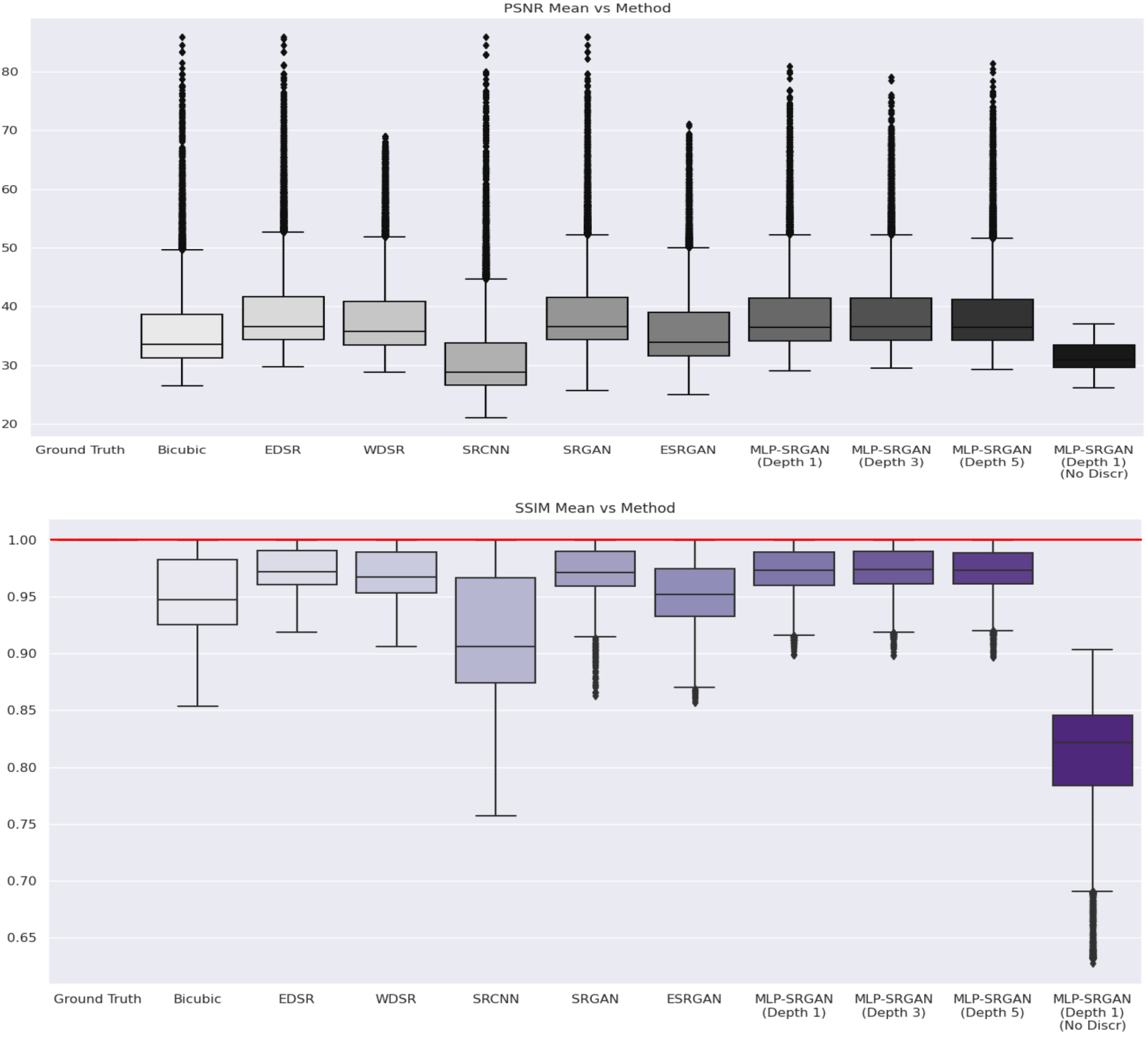}
\caption{Perceptual performance metrics from 5-fold cross validation for MSSEG2. The solid red line indicates ground truth metric. Note: ideal value for PSNR is infinity.}
\label{ref_box_fig}
\end{figure}
\begin{figure}[!ht]
\centering
\includegraphics[width=\textwidth]{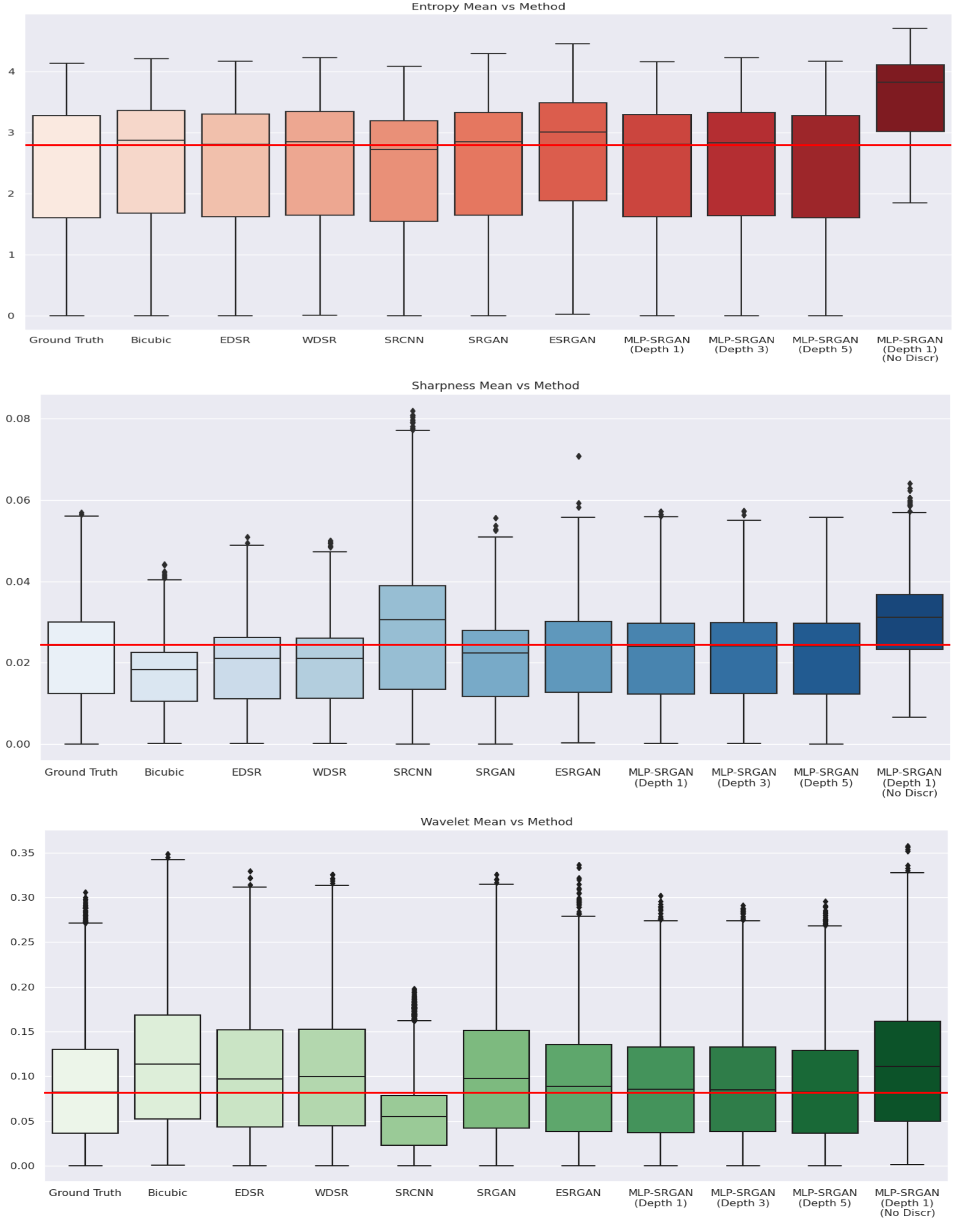}
\caption{Structural performance metrics from 5-fold cross validation for MSSEG2. The solid red line indicates ground truth metric.}
\label{no_ref_box_fig}
\end{figure}
\begin{figure}[!ht]
\centering
\includegraphics[scale=0.23]{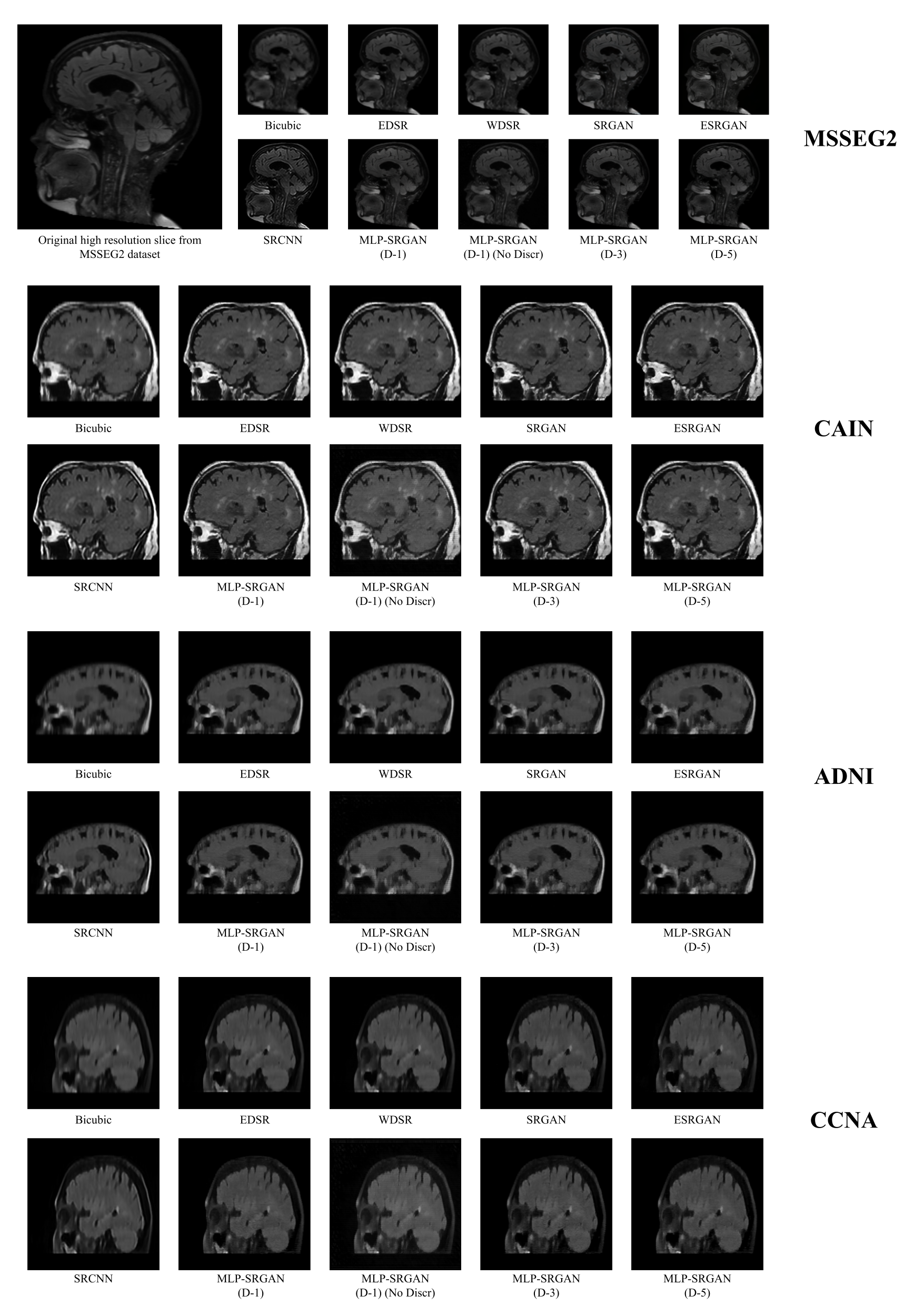}
\caption{SR results of proposed network (MLP-SRGAN). Images best viewed digitally.}
\label{results_fig}
\end{figure}
\begin{figure}[!ht]
\centering
\includegraphics[width=\textwidth]{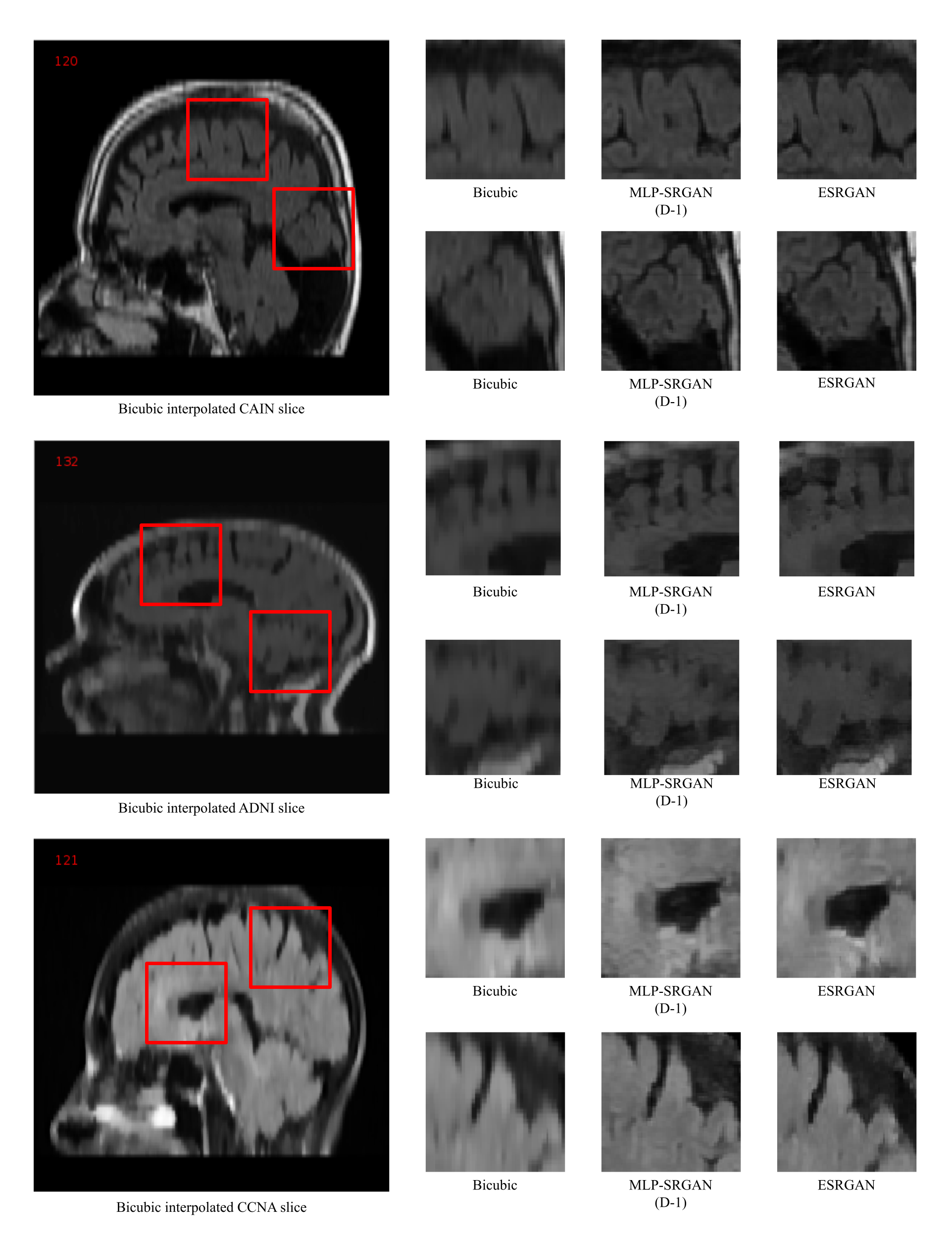}
\vspace{-10mm}
\caption{Close up images of MLP-SRGAN (D-1) results compared to bicubic interpolation and ESRGAN from the blind datasets.}
\label{results2_fig}
\end{figure}

\end{document}